\documentclass[letterpaper]{article}
\usepackage{aaai2026}
\usepackage{times}
\usepackage{helvet}
\usepackage{courier}
\usepackage[hyphens]{url}
\usepackage{graphicx}
\usepackage{natbib}
\usepackage{caption}
\usepackage{algorithm}
\usepackage{amsthm}
\usepackage{amsmath}
\usepackage{graphicx}
\usepackage{xcolor}
\usepackage{color}
\usepackage{amssymb}
\usepackage{algpseudocode}
\usepackage{mathtools}
\usepackage{epstopdf}

\usepackage{booktabs}
\usepackage{tabularx}
\usepackage{multirow}
\usepackage{adjustbox}
\usepackage{caption}
\usepackage{tabularray}
\usepackage{graphicx} 

\usepackage{array} 
\usepackage{amsmath}
\usepackage{amssymb}
\usepackage{wasysym}
\usepackage{pifont}
\usepackage{comment}
\usepackage{marvosym}

\usepackage{enumitem}

\usepackage{newfloat}
\usepackage{listings}
\DeclareCaptionStyle{ruled}{labelfont=normalfont,labelsep=colon,strut=off} 
\floatstyle{ruled}
\newfloat{listing}{tb}{lst}{}
\floatname{listing}{Listing}
\title{KG-EDAS: A Meta-Metric Framework for Evaluating Knowledge Graph Completion Models}
\author{
    Haji Gul\textsuperscript{\rm 1},
    Abul Ghani Naim\textsuperscript{\rm 1},
    Ajaz Ahmad Bhat\textsuperscript{\rm 1 $^*$}
}
\affiliations{
    \textsuperscript{\rm 1}School of Digital Science, Universiti Brunei Darussalam\\
    {(23h1710, ghani.naim, ajaz.bhat$^*$)@ubd.edu.bn}
}

\usepackage{bibentry}

\begin{document}

\maketitle

\begin{abstract}
Knowledge Graphs (KGs) enable applications in various domains such as semantic search, recommendation systems, and natural language processing. KGs are often incomplete, missing entities and relations, an issue addressed by Knowledge Graph Completion (KGC) methods that predict missing elements. Different evaluation metrics, such as Mean Reciprocal Rank (MRR), Mean Rank (MR), and Hit@k (e.g., Hit@1), are commonly used to assess the performance of such KGC models. A major challenge in evaluating KGC models however, lies in comparing their performance across multiple datasets and metrics. A model may outperform others on one dataset but underperform on another, making it difficult to determine overall superiority. Moreover, even within a single dataset, different metrics such as MRR and Hit@1 can yield conflicting rankings, where one model excels in MRR while another performs better in Hit@1, further complicating model selection for downstream tasks. These inconsistencies hinder holistic comparisons and highlight the need for a unified meta-metric that integrates performance across all metrics and datasets to enable a more reliable and interpretable evaluation framework. To address this need, we propose KG \textit{E}valuation based on \textit{D}istance from \textit{A}verage \textit{S}olution (EDAS), a robust and interpretable meta-metric that synthesizes model performance across multiple datasets and diverse evaluation criteria into a single normalized score ($M_i \in [0,1]$). Unlike traditional metrics that focus on isolated aspects of performance, EDAS offers a global perspective that supports more informed model selection and promotes fairness in cross-dataset evaluation. Experimental results on benchmark datasets such as FB15k-237 and WN18RR demonstrate that EDAS effectively integrates multi-metric, multi-dataset performance into a unified ranking, offering a consistent, robust, and generalizable framework for evaluating KGC models.
\end{abstract}


\section{Introduction}
KGs, formalized as $\mathcal{G} = (\mathcal{E}, \mathcal{R}, \mathcal{T})$ with entities $\mathcal{E}$, relations $\mathcal{R}$, and triples $\mathcal{T} \subseteq \mathcal{E} \times \mathcal{R} \times \mathcal{E}$ in the form $(h, r, t)$, encode structured real-world knowledge to enable applications such as question answering \citep{devlin2019bert}, recommendation systems \citep{zhuang-etal-2021-robustly}, and knowledge-enhanced language models. Due to their inherent incompleteness, KGC is essential for predicting missing triples, such as relation prediction given $(h, ?, t)$, tail entity $t$ prediction given $(h, r, ?)$ or head entity $h$ given $(?, r, t)$, using a scoring function $S_i: \mathcal{E} \times \mathcal{R} \times \mathcal{E} \rightarrow \mathbb{R}$ for each model $M_i$ \citep{shu2024knowledge,10.1007/978-981-96-8298-0_1}. 

Evaluating KGC models presents significant challenges, particularly when comparing their performance across multiple datasets and metrics. Commonly used rank-based metrics, such as MRR, MR, and Hits@$k$ (e.g., Hits@1, Hits@3, Hits@10), assess different aspects of model performance. However, a model may excel on one dataset while underperforming on another, making it difficult to determine overall superiority\citep{rossi2021knowledge}. Additionally, even within a single dataset, conflicting rankings often arise when different metrics are considered. For instance, a model may achieve a high MRR but a low Hits@1 score, complicating model selection and leading to inconsistent evaluations\citep{sun2020benchmarking}. These inconsistencies across datasets and metrics highlight the need for a unified meta-metric that integrates performance across diverse evaluation criteria and benchmarks to provide a comprehensive and reliable assessment of KGC models.
To address this need, we propose KG-EDAS, a multi-criteria decision-making meta-metric framework adapted from operational research~\citep{Ghorabaee2015MultiCriteriaIC} for KGC evaluation. KG-EDAS offers the following key capabilities:
\begin{itemize}    
\item \textbf{Unified Evaluation Framework:} EDAS is the first multi-criteria evaluation metric for KGC, synthesizing performance across any KGC metrics like Hits@$k$ and MR into a single normalized score $M_i \in [0,1]$ across any datasets, offering a single measure for model comparison.    
\item \textbf{Enhanced Interpretability and Robustness:} By balancing positive and negative deviations from average performance, EDAS provides interpretable global ranks (e.g., Rank 1, Rank 2) that resolve inconsistencies and reflect clear performance trade-offs.    
\item \textbf{Cross-Dataset Generalizability:} Unlike traditional metrics limited to single-dataset evaluations, EDAS enables comparisons both within and across datasets, facilitating a clearer assessment of model generalization and supporting robust model selection across benchmarks like FB15k-237 and WN18RR.    
\item \textbf{Computational Efficiency:} EDAS is implemented with linear time complexity $\mathcal{O}(nm)$, where $n$ is the number of models and $m$ is the number of evaluation criteria, ensuring scalability for large-scale KGC evaluations.
\end{itemize}
This work contributes a perspective shift in KGC evaluation by introducing a meta-metric that supports robust, interpretable and holistic comparisons of models across diverse benchmarks. Experimental results demonstrate that KG-EDAS effectively integrates multi-metric, multi-dataset performance into a unified ranking that is in consistent alignment with individual traditional metrics like MRR, MR and Hit@1 etc.
\section{Related Work}
Recent efforts in KGC have mainly focused on improving model accuracy through advanced architectures rather than refining the underlying evaluation methodologies. For instance, Wang et al.~\citep{10115028} introduced the Triplet Distributor Network (TDN), which demonstrated strong performance on Hits@3 but continued to rely on disjointed metrics such as MRR and Hits@$k$ for evaluation. Similarly, Lin et al.~\citep{lin-etal-2018-multi} proposed a multi-hop reasoning framework that achieved high MRR scores yet underperformed in Hit@1 evaluations, underscoring the inconsistent behaviour of traditional metrics across different criteria. Multi-task learning approaches, such as those by Kim et al.~\citep{kim-etal-2020-multi}, aim to enhance predictive power by integrating auxiliary tasks like relation prediction; however, they still report results using isolated metrics without addressing the broader issue of metric fragmentation. Similarly, Wei et al.~\citep{wei-etal-2023-kicgpt} introduced KICGPT, a large language model tailored for KGC, achieving competitive performance across multiple datasets. However, their evaluation strategy remains split across MRR, Hit@1, and Hit@10, requiring manual interpretation and potentially influencing comparative rankings. These examples illustrate a persistent reliance on conventional evaluation metrics despite growing recognition of their limitations. This fragmented approach complicates model comparison and hinders progress in the field, as researchers must manually weigh conflicting metric outcomes to make informed decisions.

In response to these challenges, recent studies have explored alternative strategies for evaluating KGC models. Ruffinelli et al.~\citep{ruffinelli2020you} conducted an extensive empirical review of knowledge graph embedding (KGE) models, highlighting inconsistencies in metric usage and calling for more standardized benchmarks. Sun et al.~\citep{sun2020benchmarking} emphasized the importance of incorporating uncertainty quantification into KGC evaluation, arguing that confidence estimates are essential for real-world deployment. Despite these insights, no comprehensive framework has emerged that integrates performance across multiple metrics and datasets into a single, interpretable score. Various critical gaps remain unaddressed in current KGC evaluation practices:

\textit{\textbf{(1) Lack of Cross-Dataset Comparability:}} Most evaluation frameworks are limited to single-dataset analysis, offering no mechanism to assess generalization across diverse benchmarks. As shown in recent works such as Sim-KGC~\citep{wang2022simkgc}, this limitation prevents meaningful comparisons of model robustness across varying data distributions.
\textit{\textbf{(2) Underutilization of Decision Theory:}} Although Multi Criteria Decision-Making (MCDM) methods like TOPSIS and VIKOR have demonstrated success in other machine learning domains~\citep{kandakoglu2024use}, their adoption in KGC remains minimal. These frameworks offer structured, principled ways to resolve conflicts among competing metrics and produce holistic model rankings, an opportunity largely overlooked in current KGC research.

To address these deficiencies, we introduce in the KG area a meta-meric KG-EDAS methodology derived from operational research for KGC evaluation. Unlike traditional scalar metrics such as MRR or Hit@k, which provide partial and often conflicting perspectives, EDAS synthesizes performance across multiple criteria and datasets into a unified, normalized score ($M_i \in [0,1]$). It computes both positive and negative deviations from average performance, enabling a balanced view of model strengths and weaknesses without relying on subjective reference points. This makes EDAS particularly well-suited for complex and uncertain environments like KGC, where ground truth rankings may be ambiguous or inconsistent. By introducing EDAS into the KGC domain, we present a principled, scalable, and interpretable meta-metric that supports fair and reproducible model comparison across diverse benchmarks.

\section{Methodology}
This section presents the KG-EDAS, a multi-criteria decision-making meta-metric framework for evaluating KGC models. By assessing performance across multiple metrics and datasets into a single interpretable score, KG-EDAS addresses the limitations of traditional scalar metrics like MRR and Hit@k, offering a unified and reproducible framework for model comparison. We begin by formulating the problem, followed by a structured explanation of how KG-EDAS is adapted to KGC evaluation.

\textbf{Problem Formulation:} Given a knowledge graph $G = (E, R, T)$, where $E$ denotes entities, $R$ relations, and $T \subseteq E \times R \times E$ valid triples in the form $(h, r, t)$, we focus on evaluating KGC models that predict missing entities or relations using a scoring function $S_i: E \times R \times E \rightarrow \mathbb{R}$. Let $\mathcal{M} = \{M_1, M_2, \ldots, M_n\}$ denote $n$ KGC models, each producing a vector of scores across multiple evaluation metrics such as MRR, Hit@1, Hit@10, and MR given in Equations \ref{eq:eva} and \ref{eq:evab}. 
\begin{equation}
\label{eq:eva}
\text{MR} = \frac{1}{N} \sum_{i=1}^{N} \text{rank}_i, \quad
\text{MRR} = \frac{1}{N} \sum_{i=1}^{N} \frac{1}{\text{rank}_i}, 
\end{equation}
\begin{equation}
\label{eq:evab}
 \quad   \text{Hits@k} = \frac{1}{N} \sum_{i=1}^{N} \mathbf{1}(\text{rank}_i \leq k)
\end{equation}
This study aims to derive a unified ranking of KGC models based on their aggregated performance across all evaluation metrics and datasets. To enable a holistic comparison, the process begins with the construction of a performance matrix $X \in \mathbb{R}^{n \times m}$, where each entry $X_{ij}$ represents the score of model $M_i$ on metric $j$. The following sections provide a step-by-step description of the KG-EDAS meta-metric framework.
\begin{itemize}
    \item  \textbf{Decision Matrix Construction:} To perform a systematic and multi-criteria evaluation, we organize the results into a structured format called the decision matrix $X \in \mathbb{R}^{n \times m}$, where rows correspond to models and columns to metrics:
\begin{equation}
    X =
\begin{bmatrix}
X_{11} & X_{12} & \cdots & X_{1m} \\
X_{21} & X_{22} & \cdots & X_{2m} \\
\vdots & \vdots & \ddots & \vdots \\
X_{n1} & X_{n2} & \cdots & X_{nm} \\
\end{bmatrix}
\end{equation}
$, \quad \text{where } X_{ij} \in \mathbb{R}, \; i = 1, \ldots, n, \; j = 1, \ldots, m$
Each metric is classified as either beneficial (e.g., MRR, Hit@k) or non-beneficial (e.g., MR). This matrix serves as the foundational input for the EDAS method, transforming heterogeneous performance indicators into a uniform space suitable for computing deviations from average performance.
 \item \textbf{Average Solution Computation:} Next compute the average solution $\text{Avg}_j$ for each metric $j$ as:
\begin{equation}
    \text{Avg}_j = \frac{1}{n} \sum_{i=1}^{n} X_{ij}, \quad j = 1, \ldots, m
\end{equation}
This yields an average vector $\text{Avg} = [\text{Avg}_1, \text{Avg}_2, \ldots, \text{Avg}_m] \in \mathbb{R}^m$, representing the central tendency of model performance across all criteria.
The mean solution serves as an index for evaluation, which measures the relative performance of a specific model against the rest of the group.  It provides consistency in ranking by removing biases caused by metrics using different scales; for instance, MRR generally spans from 0 to 1. The EDAS method provides a balanced and interpretable framework for multi-criteria KGC evaluation by normalising variations from the average evaluation.
 \item \textbf{Positive and Negative Distance from Average (PDA and NDA):} The next step involves measuring how each model deviates from the average solution, either positively or negatively, depending on whether the metric is beneficial or non-beneficial. This dual-metric approach helps EDAS to effectively evaluate both the strengths and weaknesses of KGC models, therefore enabling a balanced and interpretable multi-criteria ranking. Let \( X_{ij} \) as the performance score of the \( i \)-th model on the \( j \)-th criterion, and let \( \text{Avg}_j \) represent the average score of the \( j \)-th criterion across all models. 
 
For \textbf{\textit{beneficial metrics}} (e.g., MRR, Hit@k):
\begin{equation}
    \text{PDA}_{ij} = \frac{\max(0, X_{ij} - \text{Avg}_j)}{\text{Avg}_j},
\end{equation}
\begin{equation}
\quad \text{NDA}_{ij} = \frac{\max(0, \text{Avg}_j - X_{ij})}{\text{Avg}_j}
\end{equation}
For \textbf{\textit{non-beneficial metrics}} (e.g., MR):
\begin{equation}
    \text{PDA}_{ij} = \frac{\max(0, \text{Avg}_j - X_{ij})}{\text{Avg}_j}, 
\end{equation}
\begin{equation}
    \quad \text{NDA}_{ij} = \frac{\max(0, X_{ij} - \text{Avg}_j)}{\text{Avg}_j}
\end{equation}
These normalized deviations ensure comparability across metrics with different scales, avoiding division-by-zero issues via small constant adjustments where necessary.
 \item \textbf{Weighted PDA and NDA:} To incorporate the relative importance of each metric, we apply weighted aggregation. Let $w_j \in [0,1]$ denote the weight assigned to metric $j$, with $\sum_{j=1}^{m} w_j = 1$. In our experiments, an equal weight is assigned. The weighted positive and negative distances are computed as:
\begin{equation}
    \text{WPDA}_i = \sum_{j=1}^{m} w_j \cdot \text{PDA}_{ij}
\end{equation}
\begin{equation}
\quad \text{WNDA}_i = \sum_{j=1}^{m} w_j \cdot \text{NDA}_{ij}
\end{equation}
These values reflect how much better or worse a model performs relative to the average, weighted by the importance of each metric. As reported, this study used equal weights for each criterion to ensure balanced evaluation across complexity measures. However, depending on the downstream tasks such as recommendation systems or ranking applications, where metrics like Hit@10 or Hit@K are more valuable, users can assign higher weights to metrics that align with these objectives while assigning lower weights to less relevant metrics, to better tailor the KG-EDAS framework to task-specific requirements.

 \item \textbf{Normalization of WPDA and WNDA:} To ensure consistency and interpretability, we normalize WPDA and WNDA values to the range $[0,1]$:
\begin{equation}
    N(\text{WPDA}_i) = \frac{\text{WPDA}_i}{\max(\text{WPDA})}
\end{equation}
\begin{equation}
\quad N(\text{WNDA}_i) = \frac{\text{WNDA}_i}{\max(\text{WNDA})}
\end{equation}

This normalization enables meaningful comparison across diverse benchmarks.
 \item \textbf{Final Evaluation Score ($M_i$):} This step computes a unified performance score $M_i \in [0,1]$ for each model:
\begin{equation}
    M_i = \frac{1}{2} \left[ N(\text{WPDA}_i) + (1 - N(\text{WNDA}_i)) \right]
\end{equation}
This score balances strengths (positive deviation) and weaknesses (negative deviation), producing a single interpretable value for each model.
\begin{table*}[h!]
\centering
\captionsetup{justification=raggedright,singlelinecheck=false} 
\caption{Comparative time and space complexity of multi-criteria ranking methods}
\label{edas:t}
\renewcommand{\arraystretch}{1.5} 
\resizebox{\textwidth}{!}{%
\begin{tabular}{lcccc}
\hline
\textbf{Method} & \textbf{Time Complexity} & \textbf{Space Complexity} & \textbf{Parallelizable} & \textbf{Notes}  \\
\hline
\textbf{EDAS} & \( \mathcal{O}(nm) \) & \( \mathcal{O}(nm) \) & Yes & Linear in models \(\times\) metrics  \\
\textbf{TOPSIS} \citep{kandakoglu2024use}  & \( \mathcal{O}(nm + n^2) \) & \( \mathcal{O}(nm + n^2) \) & Partially & Ideal/anti-ideal vector comparisons  \\
\textbf{Pareto Frontier} \citep{lin2023pareto} & \( \mathcal{O}(n^2 m) \) & \( \mathcal{O}(n^2) \) & No & Regret Minimisation Step \\
\textbf{Borda Count} \citep{emerson2023borda}  & \( \mathcal{O}(nm \log m) \) & \( \mathcal{O}(nm) \) & Yes & Risk of inconsistency \\
\hline
\end{tabular}
}
\end{table*}

 \item \textbf{Model Ranking Based on $M_i$:} Once all models have been assigned their respective \( M_i \) scores, the final step involves generating a definitive ranking of the models based on these scores. Let \( \mathbf{M} = [M_1, M_2, \dots, M_n] \) be the vector of final scores for \( n \) models. The ranking is determined by sorting this vector in descending order:
\begin{equation}
    \text{Rank}(i) = \text{argsort}(M_i, \text{descending=True})
\end{equation}
\end{itemize}
This yields an ordered list where the model with the highest \( M_i \) receives \textit{Rank 1}, indicating superior performance across all criteria. Unlike traditional metrics such as MRR and Hit@k, which often produce conflicting rankings, the \( M_i \)-based ranking resolves inconsistencies by integrating multiple criteria into a single decision-making framework. This ranking mechanism enhances interpretability, supports fair comparison, and facilitates model selection in KGC research. 

\textbf{Computational Complexity of KG-EDAS:} As a meta-metric framework, KG-EDAS synthesizes diverse evaluation metrics—such as MRR, Hit@1, Hit@10, and MR—into a unified score $M_i \in [0,1]$, enabling holistic and interpretable comparisons across models and datasets.
One of the key strengths of EDAS lies in its linear time complexity, which ensures scalability even when evaluating large sets of models over multiple benchmark datasets. This is particularly important given the fragmented nature of KGC evaluation, where models often exhibit inconsistent performance across different metrics and datasets. Traditional scalar metrics like MRR or Hit@k are fast to compute individually but fail to provide a comprehensive view of model effectiveness. Comparing results across these traditional metrics introduces ambiguity, requiring manual inspection that becomes increasingly impractical as the number of models and evaluation criteria grows.

KG-EDAS addresses this challenge by computing a single, interpretable ranking through a structured workflow, as summarized in Table~\ref{edas:t}. Unlike more complex multi-criteria methods such as TOPSIS or VIKOR—which rely on ideal reference points or pairwise distance matrices—EDAS uses the average performance vector as a baseline, eliminating unnecessary computational overhead while maintaining robustness and fairness.

Let $n$ be the number of models being evaluated and $m$ be the number of performance criteria (e.g., MRR, Hit@1, MR). Each model's performance is represented as a row in the decision matrix $X \in \mathbb{R}^{n \times m}$. The computational steps and their respective complexities are detailed below. Let $T(n,m)$ denote the total time complexity for $n$ models and $m$ metrics. We now analyze the computational complexity of each step in the EDAS workflow:

\begin{equation}
\text{\textit{Step 1}: Average Calculation} ~ ~ ~ ~ \text{Avg}_j = \frac{1}{n} \sum_{i=1}^{n} X_{ij}
\end{equation}
\begin{equation*}
T_1(n, m) = m \cdot \mathcal{O}(n) = \mathcal{O}(nm)
\end{equation*}
\begin{equation}
\text{\textit{Step 2}: Distance Metrics} ~ ~ ~ ~  \text{PDA}_{ij}, \quad \text{NDA}_{ij}
\end{equation}
\begin{equation*}
T_2(n, m) = 2nm \cdot \mathcal{O}(1) = \mathcal{O}(nm)
\end{equation*}
\begin{equation}
\text{\textit{Step 3}: Weighted Aggregation} ~ ~ ~ ~  \text{WPDA}_i, \quad \text{WNDA}_i 
\end{equation}
\begin{equation*}
T_3(n, m) = 2n \cdot \mathcal{O}(m) = \mathcal{O}(nm)
\end{equation*}
\begin{equation}
\text{\textit{Step 4}: Normalization} ~ ~ ~ ~  N(\text{WPav}_i) , \quad N(\text{WNav}_i)
\end{equation}
\begin{equation*}
T_4(n) = \mathcal{O}(n) \text{ (max)} + \mathcal{O}(n) \text{ (division)} = \mathcal{O}(n)
\end{equation*}
\begin{equation}
\text{\textit{Step 5}: Ranking} ~ ~ ~ ~  M_i = \frac{1}{2} \left[ N(\text{WPav}_i) + (1 - N(\text{WNav}_i)) \right]
\end{equation}
\begin{equation}
\text{Rank}(i) = \text{argsort}(M_i, \text{descending=True})
\end{equation}
\begin{equation*}
T_5(n) = \mathcal{O}(n \log n)
\end{equation*}
\begin{equation}
\begin{aligned}
\textbf{\textit{Overall:}} \quad 
T(n, m) &= \underbrace{\mathcal{O}(nm)}_{\text{Steps 1--3}} 
+ \underbrace{\mathcal{O}(n)}_{\text{Step 4}} 
+ \underbrace{\mathcal{O}(n \log n)}_{\text{Step 5}} \\
&= \mathcal{O}(nm + n \log n)
\end{aligned}
\end{equation}

For typical KGC evaluations where \( m \geq \log n \) (e.g.\( n = 10^4, m = 10 \)), this simplifies to:
\begin{equation}
T(n, m) \approx \mathcal{O}(nm)
\end{equation}
\\
This linear complexity makes EDAS highly suitable for real-world applications involving large-scale model comparisons. It avoids computationally intensive operations such as iterative optimization or pairwise comparisons, further enhancing its efficiency and interpretability.

\begin{table*}[h!]
\centering
\captionsetup{justification=raggedright,singlelinecheck=false}
\caption{Relation Prediction Final EDAS Scores with Model Ranking}
\label{tab:f1}
\renewcommand{\arraystretch}{1.0}
\resizebox{\textwidth}{!}{%
\begin{tabular}{l c c c c c c}
\toprule
\textbf{Model} & \textbf{WPDA\_sum} & \textbf{WNDA\_sum} & \textbf{NWPDA} & \textbf{NWNDA} & \textbf{M} & \textbf{Rank} \\
\midrule
RotatE~\citep{sun2019rotate}       & 0.2214 & 0.0000 & 0.9954 & 0.0000 & \textbf{0.9977} & \textbf{1} \\
TuckER~\citep{16}                  & 0.1943 & 0.0075 & 0.8735 & 0.0236 & \underline{0.9250} & \underline{2} \\
RSN~\citep{jiang2019adaptive}      & 0.1590 & 0.0021 & 0.7151 & 0.0065 & 0.8543 & 3 \\
ConvR~\citep{guo2019learning}      & 0.1456 & 0.0088 & 0.6547 & 0.0277 & 0.8135 & 4 \\
ConvE~\citep{Dettmers2018ConvE}    & 0.1130 & 0.0051 & 0.5080 & 0.0158 & 0.7461 & 5 \\
DistMult~\citep{yang2015distmult}  & 0.1052 & 0.0368 & 0.4730 & 0.1155 & 0.6788 & 6 \\
CrossE~\citep{zhang2019interaction} & 0.0439 & 0.0306 & 0.1974 & 0.0960 & 0.5507 & 7 \\
SimplE~\citep{kazemi2018simple}    & 0.0333 & 0.1758 & 0.1496 & 0.5511 & 0.2992 & 8 \\
ANALOGY~\citep{liu2017analogical}  & 0.0312 & 0.1995 & 0.1404 & 0.6252 & 0.2576 & 9 \\
TorusE~\citep{ebisu2018toruse}     & 0.0718 & 0.2727 & 0.3227 & 0.8549 & 0.2339 & 10 \\
\bottomrule
\end{tabular}
}
\end{table*}
\begin{table*}[h!]
\newcommand{\numval}[1]{\normalsize #1}
\centering
\caption{Link prediction results on FB15k, WN18, FB15k-237, WN18RR, and YAGO3-10. The results reported here are published in \citep{10.1145/3424672}}
\label{tab:link_prediction_all}
\resizebox{\textwidth}{!}{%
\renewcommand{\arraystretch}{1.2}
\begin{tabular}{lcccccccccccccccccccccr}
\toprule
\multirow{2}{*}{\textbf{Models}} & \multicolumn{4}{c}{\textbf{FB15k}} 
& \multicolumn{4}{c}{\textbf{WN18}} 
& \multicolumn{4}{c}{\textbf{FB15k-237}} 
& \multicolumn{4}{c}{\textbf{WN18RR}} 
& \multicolumn{4}{c}{\textbf{YAGO3-10}} 
& \textbf{M} & \textbf{Ranks} \\
\cmidrule(lr){2-5} \cmidrule(lr){6-9} \cmidrule(lr){10-13} \cmidrule(lr){14-17} \cmidrule(lr){18-21}
& \textbf{MR} & \textbf{MRR} & \textbf{H@1} & \textbf{H@10} & \textbf{MR} & \textbf{MRR} & \textbf{H@1} & \textbf{H@10} & \textbf{MR} & \textbf{MRR} & \textbf{H@1} & \textbf{H@10} & \textbf{MR} & \textbf{MRR} & \textbf{H@1} & \textbf{H@10} & \textbf{MR} & \textbf{MRR} & \textbf{H@1} & \textbf{H@10} & & \\
\midrule
RotatE & \numval{42} & \numval{0.791} & \numval{0.739} & \numval{0.881} & \numval{274} & \numval{0.949} & \numval{0.943} & \numval{0.960} & \numval{178} & \numval{0.336} & \numval{0.238} & \numval{0.531} & \numval{3318} & \numval{0.475} & \numval{0.426} & \numval{0.573} & \numval{1827} & \numval{0.498} & \numval{0.405} & \numval{0.671} & \numval{0.998} & \textbf{1} \\
TuckER & \numval{39} & \numval{0.788} & \numval{0.729} & \numval{0.889} & \numval{510} & \numval{0.951} & \numval{0.946} & \numval{0.958} & \numval{162} & \numval{0.352} & \numval{0.259} & \numval{0.536} & \numval{6239} & \numval{0.459} & \numval{0.430} & \numval{0.514} & \numval{2417} & \numval{0.544} & \numval{0.466} & \numval{0.681} & \numval{0.925} & \underline{2} \\
RSN & \numval{70} & \numval{0.773} & \numval{0.706} & \numval{0.886} & \numval{471} & \numval{0.950} & \numval{0.946} & \numval{0.959} & \numval{251} & \numval{0.346} & \numval{0.256} & \numval{0.526} & \numval{5646} & \numval{0.467} & \numval{0.437} & \numval{0.527} & \numval{2582} & \numval{0.527} & \numval{0.446} & \numval{0.673} & \numval{0.854} & 3 \\
ConvR & \numval{51} & \numval{0.777} & \numval{0.723} & \numval{0.870} & \numval{346} & \numval{0.928} & \numval{0.912} & \numval{0.951} & \numval{248} & \numval{0.280} & \numval{0.198} & \numval{0.444} & \numval{4210} & \numval{0.395} & \numval{0.346} & \numval{0.483} & \numval{1339} & \numval{0.511} & \numval{0.427} & \numval{0.664} & \numval{0.814} & 4 \\
ConvE & \numval{51} & \numval{0.688} & \numval{0.595} & \numval{0.849} & \numval{413} & \numval{0.945} & \numval{0.939} & \numval{0.957} & \numval{281} & \numval{0.305} & \numval{0.219} & \numval{0.476} & \numval{4944} & \numval{0.427} & \numval{0.390} & \numval{0.508} & \numval{2429} & \numval{0.488} & \numval{0.399} & \numval{0.658} & \numval{0.746} & 5 \\
DistMult & \numval{173} & \numval{0.784} & \numval{0.736} & \numval{0.863} & \numval{675} & \numval{0.824} & \numval{0.726} & \numval{0.946} & \numval{199} & \numval{0.313} & \numval{0.224} & \numval{0.490} & \numval{5913} & \numval{0.433} & \numval{0.397} & \numval{0.502} & \numval{1107} & \numval{0.501} & \numval{0.413} & \numval{0.661} & \numval{0.679} & 6 \\
CrossE & \numval{136} & \numval{0.702} & \numval{0.601} & \numval{0.862} & \numval{441} & \numval{0.834} & \numval{0.733} & \numval{0.950} & \numval{227} & \numval{0.298} & \numval{0.212} & \numval{0.470} & \numval{5212} & \numval{0.405} & \numval{0.381} & \numval{0.450} & \numval{3839} & \numval{0.446} & \numval{0.331} & \numval{0.654} & \numval{0.551} & 7 \\
SimplE & \numval{138} & \numval{0.726} & \numval{0.661} & \numval{0.836} & \numval{759} & \numval{0.938} & \numval{0.933} & \numval{0.946} & \numval{651} & \numval{0.179} & \numval{0.100} & \numval{0.344} & \numval{8764} & \numval{0.398} & \numval{0.383} & \numval{0.427} & \numval{2849} & \numval{0.453} & \numval{0.358} & \numval{0.632} & \numval{0.299} & 8 \\
ANALOGY & \numval{126} & \numval{0.726} & \numval{0.656} & \numval{0.837} & \numval{808} & \numval{0.934} & \numval{0.926} & \numval{0.944} & \numval{476} & \numval{0.202} & \numval{0.126} & \numval{0.354} & \numval{9266} & \numval{0.366} & \numval{0.358} & \numval{0.380} & \numval{2423} & \numval{0.283} & \numval{0.192} & \numval{0.457} & \numval{0.258} & 9 \\
TorusE & \numval{143} & \numval{0.746} & \numval{0.689} & \numval{0.840} & \numval{525} & \numval{0.947} & \numval{0.943} & \numval{0.954} & \numval{211} & \numval{0.281} & \numval{0.196} & \numval{0.447} & \numval{4873} & \numval{0.463} & \numval{0.427} & \numval{0.534} & \numval{19455} & \numval{0.342} & \numval{0.274} & \numval{0.474} & \numval{0.234} & 10 \\
\bottomrule
\end{tabular}
}
\end{table*}
\begin{figure*}[h!]
    \centering
    \caption{\footnotesize Comparison of prediction metrics across datasets. 
    The left image shows the relation $(h, ?, t)$ prediction comparison of mean \textit{MRR} and EDAS \textit{M} values across datasets: FB15k-237, FB15k, WN18, WN18RR, and YAGO3-10. 
    The right image shows comparison of mean \textit{Hit@1} and EDAS M-values.}
    \vspace{2mm} 
    \includegraphics[width=0.48\linewidth]{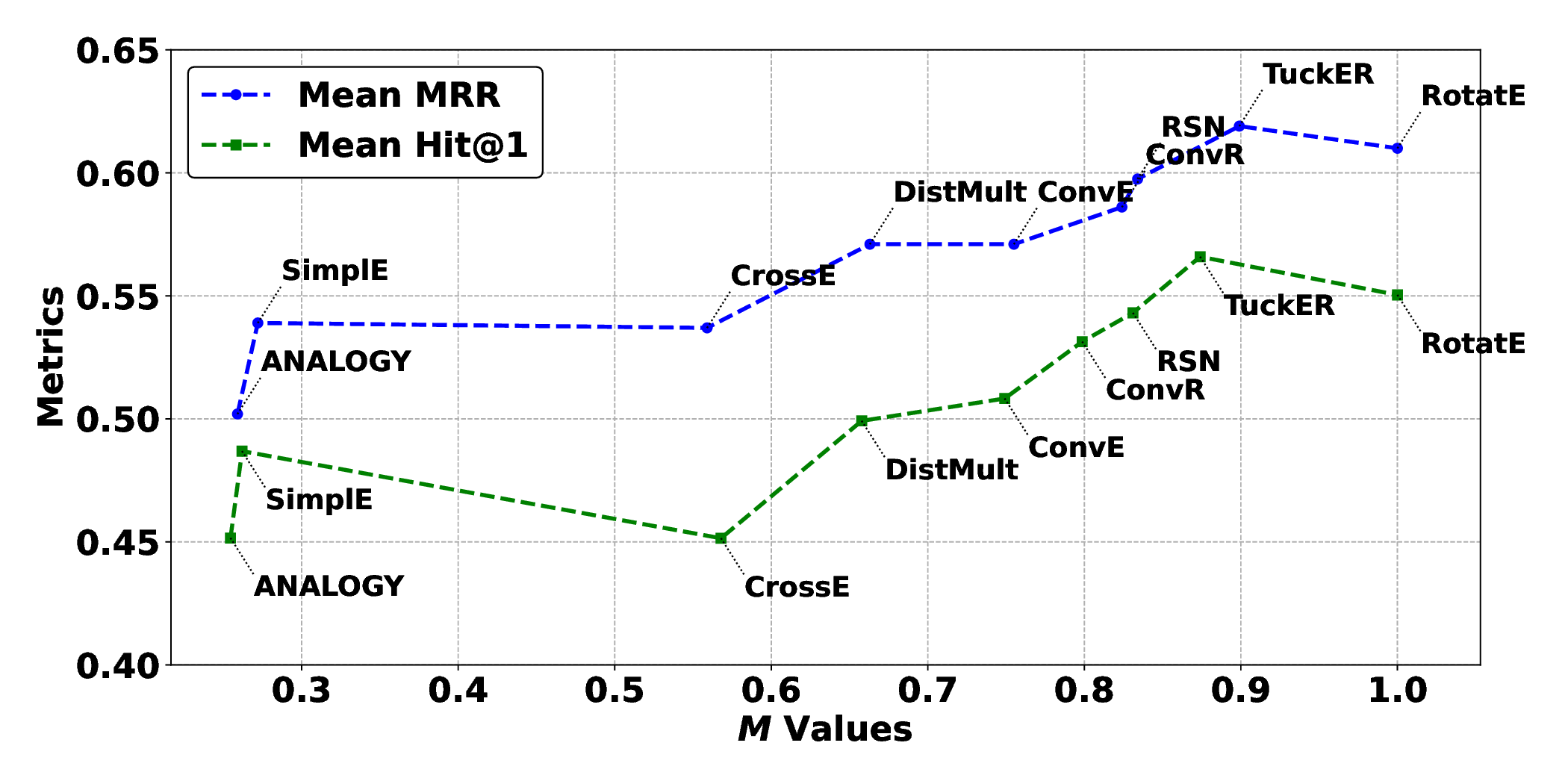}%
    \hspace{0.04\linewidth}%
    \includegraphics[width=0.48\linewidth]{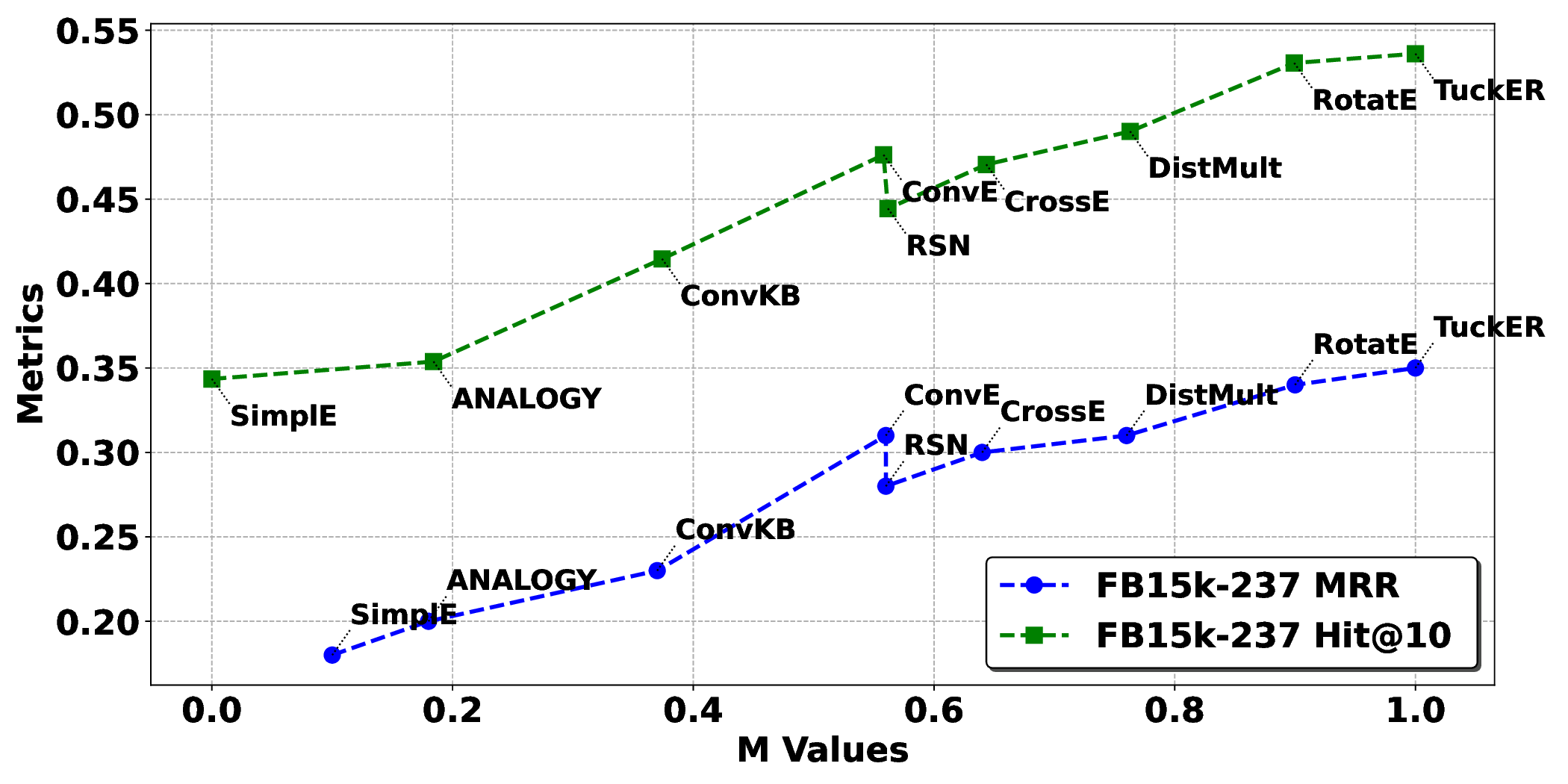}%
    \label{fig:f1}
\end{figure*}

\subsection{Experiments}
Meta-metric KG-EDAS evaluated on widely used KG datasets:
\begin{itemize}[topsep=2pt, itemsep=2pt, parsep=0pt, partopsep=0pt]
\item \textbf{YAGO3-10} \cite{mahdisoltani2013yago3}: A subset of YAGO3 focusing on high-quality facts with entities having at least 10 relations. It contains 123,182 entities, 37 relations, 1,079,040 training, 5,000 validation, and 5,000 test triplets.
\item \textbf{FB15k-237} \cite{bollacker2008freebase}: An updated version of FB15k with inverse triplets removed to increase difficulty. It consists of 14,541 entities, 237 relations, 272,115 training, 17,535 validation, and 20,466 test triplets.
\item \textbf{FB15k} \cite{bollacker2008freebase}: A subset of Freebase containing general facts. It comprises 14,951 entities, 1,345 relations, 483,142 training, 50,000 validation, and 59,071 test triplets.
\item \textbf{WN18RR} \cite{miller1995wordnet}: A subset of WN18, where reverse triplets are removed for increased complexity. The dataset includes 40,943 entities, 11 relations, 86,835 training, 2,924 validation, and 2,824 test triplets.
\item \textbf{WN18} \cite{miller1995wordnet}: A subset of WordNet with lexical relations. It includes 40,943 entities, 18 relations, 141,442 training, 5,000 validation, and 5,000 test triplets.
\end{itemize}
\section{Results}
By assessing performance across multiple metrics (MR, MRR, Hit@1, Hit@10) KG-EDAS produces a unified ranking that resolves inconsistencies often observed when using traditional metrics. 
The final EDAS score $M_i \in [0,1]$ aggregates these normalized deviations (NWPDA and NWNDA), rewarding model strengths and penalizing weaknesses in a single interpretable value. This meta-metric approach offers a holistic view of model effectiveness, addressing the limitations of traditional scalar metrics like MRR, which can be inconsistent across datasets and overly sensitive to top ranks.

\begin{table*}[h!]
\centering
\small
\captionsetup{justification=raggedright,singlelinecheck=false}
\caption{Correlation Coefficients and P-values between EDAS Score and Traditional Metrics}
\label{tab:edas_correlation}
\renewcommand{\arraystretch}{1.2}
\begin{tabularx}{\textwidth}{l l 
>{\centering\arraybackslash}X >{\centering\arraybackslash}X 
>{\centering\arraybackslash}X >{\centering\arraybackslash}X}
\toprule
\textbf{Dataset} & \textbf{Metric Pair} 
& \multicolumn{2}{c}{\textbf{Pearson}} 
& \multicolumn{2}{c}{\textbf{Kendall}} \\
\cmidrule(lr){3-4} \cmidrule(lr){5-6}
& & \textbf{Correlation} & \textbf{P-value} 
  & \textbf{Correlation} & \textbf{P-value} \\
\midrule 
\multirow{2}{*}{Multiple Datasets} 
& EDAS M Values vs Mean\_MRR    & 0.9332 & 0.0002 & 0.8733 & 0.0012 \\
& EDAS M Values vs Mean\_Hit@1  & 0.8329 & 0.0053 & 0.8333 & 0.0009 \\
\hline \\
\multirow{3}{*}{FB15k-237} 
& EDAS M Values vs Hit@10   & 0.9834 & 0.0000 & 0.8889 & 0.0002 \\
& EDAS M Values vs MRR     & 0.9739 & 0.0000 & 0.9143 & 0.0007 \\
& EDAS M Values vs MR      & -0.8372 & 0.0025 & -0.6889 & 0.0047 \\
\bottomrule
\end{tabularx}
\end{table*}
\begin{table*}[h!]
\centering
\captionsetup{justification=raggedright,singlelinecheck=false}
\renewcommand{\arraystretch}{1.0}
\caption{Tail Prediction Final EDAS Scores with Model Ranking}
\label{tab:f}
\resizebox{\textwidth}{!}{%
\begin{tabular}{l c c c c c c}
\toprule
\textbf{Model} & \textbf{WPDA\_sum} & \textbf{WNDA\_sum} & \textbf{NWPDA} & \textbf{NWNDA} & \textbf{M} & \textbf{Rank} \\
\midrule
TransR~\citep{lin2015learning}         & 0.1745 & 0.0331 & 0.9482 & 0.0604 &\textbf{0.9439} & \textbf{1} \\
TransD~\citep{ji2015knowledge}         & 0.1767 & 0.0677 & 0.9603 & 0.1236 & \underline{0.9183} & \underline{2} \\
TransH~\citep{wang2014knowledge}       & 0.1629 & 0.0619 & 0.8851 & 0.1130 & 0.8860 & 3 \\
TransE~\citep{bordes2013translating}   & 0.1362 & 0.0498 & 0.7404 & 0.0909 & 0.8248 & 4 \\
ComplEx~\citep{trouillon2016complex}   & 0.0567 & 0.0819 & 0.3080 & 0.1495 & 0.5793 & 5 \\
DistMult~\citep{yang2015distmult}      & 0.0641 & 0.1128 & 0.3485 & 0.2060 & 0.5713 & 6 \\
AMIE~\citep{galarraga2015amie}         & 0.1840 & 0.5478 & 1.0000 & 1.0000 & 0.5000 & 7 \\
\bottomrule
\end{tabular}
}
\end{table*}

\begin{table*}[h!]
\centering
\caption{Tail prediction results on FB15k, WN18, FB15k-237, and WN18RR using baseline models, including aggregated metric \( M \). The results reported here are published in \citep{akrami2020realistic}.}
\label{tab:baseline_tail_prediction_sorted}
\resizebox{\textwidth}{!}{%
\renewcommand{\arraystretch}{1.2}
\begin{tabular}{lcccccccccccccccccc}
\toprule
\textbf{Model} & \multicolumn{4}{c}{\textbf{FB15k}} & \multicolumn{4}{c}{\textbf{WN18}} & \multicolumn{4}{c}{\textbf{FB15k-237}} & \multicolumn{4}{c}{\textbf{WN18RR}} & \textbf{M} & \textbf{Ranking} \\
\cmidrule(lr){2-5} \cmidrule(lr){6-9} \cmidrule(lr){10-13} \cmidrule(lr){14-17}
 & \textbf{MR} & \textbf{MRR} & \textbf{Hits@10} & & \textbf{MR} & \textbf{MRR} & \textbf{Hits@10} & & \textbf{MR} & \textbf{MRR} & \textbf{Hits@10} & & \textbf{MR} & \textbf{MRR} & \textbf{Hits@10} & & & \\
\midrule
TransE  & {\large 243} & {\large 0.227} & {\large 0.199} & & {\large 263} & {\large 0.395} & {\large 0.142} & & {\large 363.3} & {\large 0.169} & {\large 0.32} & & {\large 2414.7} & {\large 0.176} & {\large 0.47} & & {\large 0.944} & \textbf{\large 1} \\
TransH & {\large 211} & {\large 0.177} & {\large 0.234} & & {\large 318} & {\large 0.434} & {\large 0.190} & & {\large 398.8} & {\large 0.157} & {\large 0.30} & & {\large 2616} & {\large 0.178} & {\large 0.46} & & {\large 0.918} & \underline{\large 2} \\
TransR  & {\large 226} & {\large 0.236} & {\large 0.231} & & {\large 232} & {\large 0.441} & {\large 0.199} & & {\large 391.3} & {\large 0.164} & {\large 0.31} & & {\large 2847} & {\large 0.184} & {\large 0.48} & & {\large 0.886} & \large 3 \\
TransD  & {\large 211} & {\large 0.179} & {\large 0.234} & & {\large 242} & {\large 0.421} & {\large 0.202} & & {\large 391.6} & {\large 0.154} & {\large 0.30} & & {\large 2967} & {\large 0.172} & {\large 0.47} & & {\large 0.825} & \large 4 \\
DistMult & {\large 313} & {\large 0.240} & {\large 0.264} & & {\large 915} & {\large 0.558} & {\large 0.80} & & {\large 566.3} & {\large 0.151} & {\large 0.30} & & {\large 3798.1} & {\large 0.264} & {\large 0.46} & & {\large 0.579} & \large 5 \\
ComplEx  & {\large 350.3} & {\large 0.233} & {\large 0.250} & & {\large 636.1} & {\large 0.584} & {\large 0.80} & & {\large 656.4} & {\large 0.158} & {\large 0.29} & & {\large 3755.9} & {\large 0.276} & {\large 0.46} & & {\large 0.571} & \large 6 \\
AMIE  & {\large 337} & {\large 0.370} & {\large 0.64} & & {\large 1299.8} & {\large 0.931} & {\large 0.094} & & {\large 1909} & {\large 0.201} & {\large 0.36} & & {\large 12963} & {\large 0.357} & {\large 0.35} & & {\large 0.500} & \large 7 \\
\bottomrule
\end{tabular}%
}
\end{table*}

We first apply KG-EDAS to link (relation) prediction task results from different models across different datasets as shown in Tables \ref{tab:f1} and \ref{tab:link_prediction_all}. As Table \ref{tab:f1} shows,  RotatE achieves the highest EDAS score ($M_i = 0.9977$) and is ranked first due to its consistently strong performance across all datasets, reflected in its high NWPDA (0.9954) and zero NWNDA. In contrast, models such as ANALOGY ($M_i = 0.2576$) and TorusE ($M_i = 0.2339$) receive lower rankings due to higher NWNDA values, indicating more frequent underperformance relative to the group average. The experimental results summarized in Table~\ref{tab:link_prediction_all} further demonstrate that KG-EDAS effectively resolves conflicts among conventional metrics, delivering a definitive and interpretable ranking of KGC models. This enables fair comparisons not only within individual benchmarks but also across them, supporting generalizable insights into model selection. 

Correlation Analysis of EDAS with KGC methods given in Figure~\ref{fig:f1}(a) illustrates the relationship between the proposed KG-EDAS score ($M$) and traditional evaluation metrics Mean MRR and Mean Hit@1—across multiple benchmark datasets including FB15k, WN18, FB15k-237, WN18RR, and YAGO3-10. When models are ranked by their EDAS scores along the $x$-axis, it becomes evident that both Mean MRR and Mean Hit@1 exhibit strong positive correlations with $M$, particularly in distinguishing top-performing models. This suggests that EDAS effectively captures the core strengths emphasized by these widely used metrics while resolving inconsistencies that arise when models perform well in one metric but poorly in another. In contrast, isolated scalar metrics often produce conflicting rankings, making it difficult to derive a reliable overall assessment of model performance. EDAS addresses this issue by synthesizing these metrics into a single, interpretable score, offering a more balanced and consistent evaluation framework.

Figure~\ref{fig:f1}(b), focusing on the FB15k-237 dataset, further reinforces the consistency of EDAS with conventional metrics such as Hit@10. The plot shows a clear pattern, indicating that models achieving higher Hit@10 values also receive higher EDAS scores. This graphical alignment supports the hypothesis that EDAS preserves and enhances the meaningful insights captured by individual metrics while eliminating ambiguity caused by conflicting rankings. Unlike traditional metrics that fluctuate independently and may misrepresent performance robustness, EDAS aggregates results across all criteria and datasets, producing a stable and interpretable ranking that reflects true model strength.

The statistical correlation analysis presented in Table~\ref{tab:edas_correlation} quantifies this alignment. Across multiple datasets, EDAS demonstrates a strong correlation with both Mean MRR and Mean Hit@1, with Pearson coefficient values at 0.9332 and 0.8329 respectively, both statistically significant at $p < 0.01$. Kendall’s $\tau$ confirms this strong agreement, showing values of 0.8733 and 0.8333, respectively. On the FB15k-237 dataset specifically, the correlation is even stronger, with Pearson values of 0.9834 for Hit@10 and 0.9739 for MRR. These results validate that EDAS not only aligns closely with established metrics but also enhances evaluation stability by integrating them into a unified meta-metric framework. While MR exhibits a moderate negative correlation (Pearson = $-0.8372$), this too is expected and consistent, reflecting EDAS’s ability to reward low MR values appropriately. Altogether, these findings confirm that EDAS reliably reflects model quality as assessed by traditional metrics, while offering a more holistic and reproducible evaluation approach.

Similarly, for the tail prediction task results, illustrated in Table~\ref{tab:f}, utilizing the KG-EDAS further substantiates its cross-dataset capability and unique rank allocation. Upon evaluating each method across the datasets in Table \ref{tab:baseline_tail_prediction_sorted}, it is clear that EDAS eliminates deficiencies seen in conventional metrics, providing an accurate and coherent ranking of KGC models. Moreover, the linear time complexity $\mathcal{O}(nm)$ of the EDAS method ensures scalability and efficiency, making it particularly suitable for large-scale KGC evaluations involving many models and diverse evaluation criteria. 
\section{Ablation}
To evaluate the sensitivity of KG-EDAS to individual assessment metrics and confirm its robustness, we conducted an ablation study by sequentially removing one metric at a time MRR, MR, and Hit@1—and recomputing the EDAS model rankings. The results, summarized in Table~\ref{tab:removed_metric_ranking}, demonstrate that KG-EDAS produces highly consistent rankings even when a key metric is excluded.
\begin{table}[h!]
\centering
\renewcommand{\arraystretch}{1.5}
\caption{Model Ranking Analysis After Removing Individual Metrics}
\label{tab:removed_metric_ranking}
\resizebox{\columnwidth}{!}{%
\begin{tabular}{l c c c c c}
\toprule
 & \textbf{Original} & \multicolumn{3}{c}{\textbf{Removed}} & \\
\cmidrule(lr){3-5}
\textbf{Model} & \textbf{Rank} & \textbf{MRR} & \textbf{MR} & \textbf{Hit@1} & \textbf{Max Change} \\
\midrule
RotatE     & 1 & 1 & 3 & 1 & 2 \\
TuckER     & 2 & 2 & 1 & 2 & 1 \\
ConvR      & 3 & 3 & 2 & 3 & 1 \\
ConvE      & 4 & 4 & 5 & 4 & 1 \\
DistMult   & 5 & 5 & 4 & 5 & 1 \\
CrossE     & 6 & 6 & 6 & 6 & 0 \\
SimplE     & 7 & 7 & 7 & 7 & 0 \\
ANALOGY    & 8 & 8 & 8 & 8 & 0 \\
\bottomrule
\end{tabular}%
}
\end{table}

When MRR is removed, the rankings remain identical to the original KG-EDAS ranking for all models. This indicates that MRR, while informative, does not disproportionately influence the final ranking. Similarly, excluding Hit@1 results in no rank changes across any model, confirming that the framework effectively captures performance through complementary metrics without over-reliance on top-1 accuracy.
In contrast, removing MR leads to more noticeable shifts—most notably, TuckER and ConvR swap positions, and ConvE drops from rank 4 to 5. RotatE exhibits the largest movement, shifting from rank 1 to rank 3 when MR is removed (a change of 2 positions), as reflected in its maximum rank change value. This suggests that MR plays a distinctive role in differentiating models with mid-tier performance, where subtle differences in ranking quality become more evident.
Despite these changes, the majority of models show minimal variation. In fact, three models (CrossE, SimplE, ANALOGY) maintain identical rankings across all ablation settings (Max Change = 0), and no model experiences a rank shift larger than 2 positions. This further underscores the stability of the framework.

These results confirm that KG-EDAS provides a balanced and robust evaluation: it integrates multiple performance aspects into a single score without being unduly influenced by any individual metric. This makes it a reliable and consistent alternative for evaluating and ranking knowledge graph completion models, even under partial evaluation conditions.

\section{Conclusion}
In conclusion, \textbf{KG-EDAS} is a holistic and interpretable meta-metric framework for evaluating KGC models across multiple datasets and performance criteria. By integrating both positive and negative deviations from average performance, EDAS offers a balanced view of model strengths and weaknesses, capturing trade-offs that conventional metrics miss, such as high MRR but low Hit@1.
The experimental results demonstrate that KG-EDAS aligns strongly with established metrics like mean MRR and mean Hit@1 while resolving inconsistencies among them. Correlation analysis shows that EDAS closely matches these metrics, especially MRR, while providing a stronger and more reliable way to rank results. Furthermore, ablation studies show that the framework remains largely stable even when individual metrics are removed, highlighting its resilience and comprehensive design.
By looking at more than just single metrics, KG-EDAS allows for consistent comparisons of models across different datasets and helps researchers make better decisions in KGC studies. Its linear time complexity ensures scalability, making it suitable for large-scale model assessments. These advantages position KG-EDAS as a valuable tool not only for benchmarking KGC methods but also for guiding future model development and selection.
This work brings a change in how to evaluate KGC, moving from scattered, specific metrics for each dataset to a clear and consistent framework for evaluation. As KGs continue to grow in size and application scope, such a standardized and interpretable evaluation methodology becomes essential for meaningful progress in the field.


\bigskip
\noindent Thank you for reading these instructions carefully. We look forward to receiving your electronic files!

\bibliography{aaai2026}

\begin{thebibliography}{37}
\providecommand{\natexlab}[1]{#1}

\bibitem[{Akrami et~al.(2020)Akrami, Saeef, Zhang, Hu, and Li}]{akrami2020realistic}
Akrami, F.; Saeef, M.~S.; Zhang, Q.; Hu, W.; and Li, C. 2020.
\newblock Realistic re-evaluation of knowledge graph completion methods: An experimental study.
\newblock In \emph{Proceedings of the 2020 ACM SIGMOD International Conference on Management of Data}, 1995--2010.

\bibitem[{Bollacker et~al.(2008)Bollacker, Evans, Paritosh, Sturge, and Taylor}]{bollacker2008freebase}
Bollacker, K.; Evans, C.; Paritosh, P.; Sturge, T.; and Taylor, J. 2008.
\newblock Freebase: A collaboratively created graph database for structuring human knowledge.
\newblock In \emph{Proc. of ACM SIGMOD}, 1247--1250.

\bibitem[{Bordes et~al.(2013)Bordes, Usunier, Garcia-Duran, Weston, and Yakhnenko}]{bordes2013translating}
Bordes, A.; Usunier, N.; Garcia-Duran, A.; Weston, J.; and Yakhnenko, O. 2013.
\newblock Translating embeddings for modeling multi-relational data.
\newblock \emph{Advances in neural information processing systems}, 26.

\bibitem[{Dettmers et~al.(2018)Dettmers, Minervini, Stenetorp, and Riedel}]{Dettmers2018ConvE}
Dettmers, T.; Minervini, P.; Stenetorp, P.; and Riedel, S. 2018.
\newblock Convolutional 2D knowledge graph embeddings.
\newblock In \emph{Proc. of AAAI}.

\bibitem[{Devlin et~al.(2019)Devlin, Chang, Lee, and Toutanova}]{devlin2019bert}
Devlin, J.; Chang, M.-W.; Lee, K.; and Toutanova, K. 2019.
\newblock Bert: Pre-training of deep bidirectional transformers for language understanding.
\newblock In \emph{Proceedings of the 2019 conference of the North American chapter of the association for computational linguistics: human language technologies, volume 1 (long and short papers)}, volume~1, 4171--4186.

\bibitem[{Ebisu and Ichise(2018)}]{ebisu2018toruse}
Ebisu, T.; and Ichise, R. 2018.
\newblock Toruse: Knowledge Graph Embedding on a Lie Group.
\newblock In \emph{Proceedings of the AAAI Conference on Artificial Intelligence}, volume~32.

\bibitem[{Emerson(2023)}]{emerson2023borda}
Emerson, P. 2023.
\newblock From Borda to Approval Voting: A Comparative Analysis of Decision-Making Methods.
\newblock \emph{European Journal of Operational Research}, 305(1): 1--12.

\bibitem[{Gal{\'a}rraga et~al.(2015)Gal{\'a}rraga, Teflioudi, Hose, and Suchanek}]{galarraga2015amie}
Gal{\'a}rraga, L.~A.; Teflioudi, C.; Hose, K.; and Suchanek, F.~M. 2015.
\newblock AMIE: association rule mining under incomplete evidence in ontological knowledge bases.
\newblock \emph{Proceedings of the 24th International Conference on World Wide Web}, 413--422.

\bibitem[{Ghorabaee et~al.(2015)Ghorabaee, Zavadskas, Olfat, and Turskis}]{Ghorabaee2015MultiCriteriaIC}
Ghorabaee, M.~K.; Zavadskas, E.~K.; Olfat, L.; and Turskis, Z. 2015.
\newblock Multi-Criteria Inventory Classification Using a New Method of Evaluation Based on Distance from Average Solution (EDAS).
\newblock \emph{Informatica}, 26: 435--451.

\bibitem[{Gul, Naim, and Bhat(2025)}]{10.1007/978-981-96-8298-0_1}
Gul, H.; Naim, A.~G.; and Bhat, A.~A. 2025.
\newblock MuCo-KGC: Multi-context-Aware Knowledge Graph Completion.
\newblock In Wu, X.; Spiliopoulou, M.; Wang, C.; Kumar, V.; Cao, L.; Zhou, X.; Pang, G.; and Gama, J., eds., \emph{Data Science: Foundations and Applications}, 3--15. Singapore: Springer Nature Singapore.
\newblock ISBN 978-981-96-8298-0.

\bibitem[{Guo, Sun, and Hu(2019)}]{guo2019learning}
Guo, L.; Sun, Z.; and Hu, W. 2019.
\newblock Learning to Exploit Long-term Relational Dependencies in Knowledge Graphs.
\newblock In \emph{International Conference on Machine Learning}. PMLR.

\bibitem[{Ji et~al.(2015)Ji, He, Xu, Liu, and Zhao}]{ji2015knowledge}
Ji, G.; He, S.; Xu, L.; Liu, K.; and Zhao, J. 2015.
\newblock Knowledge graph embedding via dynamic mapping matrix.
\newblock \emph{Proceedings of the 53rd Annual Meeting of the Association for Computational Linguistics and the 7th International Joint Conference on Natural Language Processing (Volume 1: Long Papers)}, 687--696.

\bibitem[{Jiang, Wang, and Wang(2019)}]{jiang2019adaptive}
Jiang, X.; Wang, Q.; and Wang, B. 2019.
\newblock Adaptive Convolution for Multi-relational Learning.
\newblock In \emph{Proceedings of the 2019 Conference of the North American Chapter of the Association for Computational Linguistics: Human Language Technologies, Volume 1 (Long and Short Papers)}.

\bibitem[{Kandakoglu, Walther, and Ben~Amor(2024)}]{kandakoglu2024use}
Kandakoglu, M.; Walther, G.; and Ben~Amor, S. 2024.
\newblock The use of multi-criteria decision-making methods in project portfolio selection: a literature review and future research directions.
\newblock \emph{Annals of Operations Research}, 332(1): 807--830.

\bibitem[{Kazemi and Poole(2018)}]{kazemi2018simple}
Kazemi, S.~M.; and Poole, D. 2018.
\newblock Simple Embedding for Link Prediction in Knowledge Graphs.
\newblock In \emph{Advances in Neural Information Processing Systems}, volume~31.

\bibitem[{Kim et~al.(2020)Kim, Hong, Ko, and Seo}]{kim-etal-2020-multi}
Kim, B.; Hong, T.; Ko, Y.; and Seo, J. 2020.
\newblock Multi-Task Learning for Knowledge Graph Completion with Pre-trained Language Models.
\newblock In Scott, D.; Bel, N.; and Zong, C., eds., \emph{Proceedings of the 28th International Conference on Computational Linguistics}, 1737--1743. Barcelona, Spain (Online): International Committee on Computational Linguistics.

\bibitem[{Lin, Zhang, and Wang(2023)}]{lin2023pareto}
Lin, X.; Zhang, Q.; and Wang, J. 2023.
\newblock Pareto Frontier Learning for Multi-Objective Optimization in Machine Learning.
\newblock \emph{IEEE Transactions on Neural Networks and Learning Systems}.

\bibitem[{Lin, Socher, and Xiong(2018)}]{lin-etal-2018-multi}
Lin, X.~V.; Socher, R.; and Xiong, C. 2018.
\newblock Multi-Hop Knowledge Graph Reasoning with Reward Shaping.
\newblock In Riloff, E.; Chiang, D.; Hockenmaier, J.; and Tsujii, J., eds., \emph{Proceedings of the 2018 Conference on Empirical Methods in Natural Language Processing}, 3243--3253. Brussels, Belgium: Association for Computational Linguistics.

\bibitem[{Lin et~al.(2015)Lin, Liu, Sun, Liu, and Zhu}]{lin2015learning}
Lin, Y.; Liu, Z.; Sun, M.; Liu, Y.; and Zhu, X. 2015.
\newblock Learning entity and relation embeddings for knowledge graph completion.
\newblock In \emph{Proceedings of the AAAI conference on artificial intelligence}, volume~29.

\bibitem[{Liu, Wu, and Yang(2017)}]{liu2017analogical}
Liu, H.; Wu, Y.; and Yang, Y. 2017.
\newblock Analogical Inference for Multi-relational Embeddings.
\newblock In \emph{International Conference on Machine Learning}. PMLR.

\bibitem[{Mahdisoltani, Biega, and Suchanek(2013)}]{mahdisoltani2013yago3}
Mahdisoltani, F.; Biega, J.; and Suchanek, F.~M. 2013.
\newblock YAGO3: A Knowledge Base from Multilingual Wikipedias.
\newblock In \emph{7th Biennial Conference on Innovative Data Systems Research (CIDR 2013)}.
\newblock Online; accessed YAGO3-10 dataset details.

\bibitem[{Miller(1995)}]{miller1995wordnet}
Miller, G.~A. 1995.
\newblock WordNet: A lexical database for English.
\newblock \emph{Communications of the ACM}, 38(11).

\bibitem[{Rossi et~al.(2021{\natexlab{a}})Rossi, Barbosa, Firmani, Matinata, and Merialdo}]{rossi2021knowledge}
Rossi, A.; Barbosa, D.; Firmani, D.; Matinata, A.; and Merialdo, P. 2021{\natexlab{a}}.
\newblock Knowledge graph embedding for link prediction: A comparative analysis.
\newblock \emph{TKDD}.

\bibitem[{Rossi et~al.(2021{\natexlab{b}})Rossi, Barbosa, Firmani, Matinata, and Merialdo}]{10.1145/3424672}
Rossi, A.; Barbosa, D.; Firmani, D.; Matinata, A.; and Merialdo, P. 2021{\natexlab{b}}.
\newblock Knowledge Graph Embedding for Link Prediction: A Comparative Analysis.
\newblock \emph{ACM Trans. Knowl. Discov. Data}, 15(2).

\bibitem[{Ruffinelli, Broscheit, and Gemulla(2020)}]{ruffinelli2020you}
Ruffinelli, D.; Broscheit, S.; and Gemulla, R. 2020.
\newblock You can teach an old dog new tricks! on training knowledge graph embeddings.

\bibitem[{Shu et~al.(2024)Shu, Chen, Jin, Zhang, Du, and Zhang}]{shu2024knowledge}
Shu, D.; Chen, T.; Jin, M.; Zhang, C.; Du, M.; and Zhang, Y. 2024.
\newblock Knowledge Graph Large Language Model (KG-LLM) for Link Prediction:(ACML).
\newblock \emph{Proceedings of Machine Learning Research}, 260(1): 143.

\bibitem[{Sun et~al.(2019)Sun, Deng, Nie, and Tang}]{sun2019rotate}
Sun, Z.; Deng, Z.; Nie, J.; and Tang, J. 2019.
\newblock RotatE: Knowledge Graph Embedding by Relational Rotation in Complex Space.
\newblock In \emph{Proc. of ICLR}.

\bibitem[{Sun et~al.(2020)Sun, Zhang, Hu, Wang, Chen, Akrami, and Li}]{sun2020benchmarking}
Sun, Z.; Zhang, Q.; Hu, W.; Wang, C.; Chen, M.; Akrami, F.; and Li, C. 2020.
\newblock A benchmarking study of embedding-based entity alignment for knowledge graphs.
\newblock \emph{arXiv preprint arXiv:2003.07743}.

\bibitem[{Trouillon et~al.(2016)Trouillon, Welbl, Riedel, Gaussier, and Bouchard}]{trouillon2016complex}
Trouillon, T.; Welbl, J.; Riedel, S.; Gaussier, {\'E}.; and Bouchard, G. 2016.
\newblock Complex embeddings for simple link prediction.
\newblock In \emph{International Conference on Machine Learning}, 2071--2080. PMLR.

\bibitem[{Wang et~al.(2023)Wang, Wang, Gao, Li, Hu, and Yin}]{10115028}
Wang, J.; Wang, B.; Gao, J.; Li, X.; Hu, Y.; and Yin, B. 2023.
\newblock TDN: Triplet Distributor Network for Knowledge Graph Completion.
\newblock \emph{IEEE Transactions on Knowledge and Data Engineering}, 35(12): 13002--13014.

\bibitem[{Wang et~al.(2022)Wang, Zhao, Wei, and Liu}]{wang2022simkgc}
Wang, L.; Zhao, W.; Wei, Z.; and Liu, J. 2022.
\newblock SIM-KGC: Simple Contrastive KGC with Pre-trained Language Models.
\newblock In \emph{Proc. of ACL}.

\bibitem[{Wang, Broscheit, and Gemulla(2019)}]{16}
Wang, Y.; Broscheit, S.; and Gemulla, R. 2019.
\newblock A Relational Tucker Decomposition for Multi-Relational Link Prediction.
\newblock \emph{arXiv preprint}.
\newblock ArXiv:1902.00898.

\bibitem[{Wang et~al.(2014)Wang, Zhang, Feng, and Chen}]{wang2014knowledge}
Wang, Z.; Zhang, J.; Feng, J.; and Chen, Z. 2014.
\newblock Knowledge graph embedding by translating on hyperplanes.
\newblock In \emph{Proceedings of the AAAI Conference on Artificial Intelligence}, volume~28.

\bibitem[{Wei et~al.(2023)Wei, Huang, Zhang, and Kwok}]{wei-etal-2023-kicgpt}
Wei, Y.; Huang, Q.; Zhang, Y.; and Kwok, J. 2023.
\newblock {KICGPT}: Large Language Model with Knowledge in Context for Knowledge Graph Completion.
\newblock In Bouamor, H.; Pino, J.; and Bali, K., eds., \emph{Findings of the Association for Computational Linguistics: EMNLP 2023}, 8667--8683. Singapore: Association for Computational Linguistics.

\bibitem[{Yang et~al.(2015)Yang, Yih, He, Gao, and Deng}]{yang2015distmult}
Yang, B.; Yih, W.-t.; He, X.; Gao, J.; and Deng, L. 2015.
\newblock Embedding entities and relations for learning and inference in knowledge bases.
\newblock \emph{arXiv preprint arXiv:1412.6575}.

\bibitem[{Zhang et~al.(2019)Zhang, Paudel, Zhang, Bernstein, and Chen}]{zhang2019interaction}
Zhang, W.; Paudel, B.; Zhang, W.; Bernstein, A.; and Chen, H. 2019.
\newblock Interaction Embeddings for Prediction and Explanation in Knowledge Graphs.
\newblock In \emph{Proceedings of the Twelfth ACM International Conference on Web Search and Data Mining}.

\bibitem[{Zhuang et~al.(2021)Zhuang, Wayne, Ya, and Jun}]{zhuang-etal-2021-robustly}
Zhuang, L.; Wayne, L.; Ya, S.; and Jun, Z. 2021.
\newblock A Robustly Optimized {BERT} Pre-training Approach with Post-training.
\newblock In Li, S.; Sun, M.; Liu, Y.; Wu, H.; Liu, K.; Che, W.; He, S.; and Rao, G., eds., \emph{Proceedings of the 20th Chinese National Conference on Computational Linguistics}, 1218--1227. Huhhot, China: Chinese Information Processing Society of China.

\end{thebibliography}

\end{document}